\documentclass[11pt,a4paper]{article}
\usepackage{authblk}
\usepackage[hyperref]{eacl2021}
\usepackage{times}
\usepackage{latexsym}

\usepackage{makecell}
\usepackage{framed}
\usepackage{graphicx}
\graphicspath{{./}}
\usepackage{amsfonts} 
\usepackage{amsmath}
\usepackage{multirow}
\usepackage{siunitx}

\usepackage{adjustbox}
\usepackage{microtype}
\usepackage{dirtytalk}
\usepackage{tabularx,ragged2e}
\newcolumntype{C}{>{\Centering\arraybackslash}X} 

\aclfinalcopy 

\title{PVG at WASSA 2021: A Multi-Input, Multi-Task, Transformer-Based Architecture for Empathy and Distress Prediction}

\author[1]{\textbf{Atharva Kulkarni}}
\author[1]{\textbf{Sunanda Somwase}}
\author[1]{\textbf{Shivam Rajput}}
\author[2]{\textbf{Manisha Marathe}}

\affil[1, 2]{Department of Computer Engineering, PVG's College of Engineering and Technology, \protect\\
affiliated to Savitribai Phule Pune University, India.}

\affil[1]{\texttt {{\fontsize{11}{6}\selectfont \{atharva.j.kulkarni1998, sunandasomwase, shivamraje43\}}\fontsize{11}{6}\selectfont @gmail.com}}
\affil[2]{\texttt { \fontsize{11}{6}\selectfont mvm\_comp@pvgcoet.ac.in }}

\date{}

\begin{document}
\maketitle
\begin{abstract}
Active research pertaining to the affective phenomenon of empathy and distress is invaluable for improving human-machine interaction. Predicting intensities of such complex emotions from textual data is difficult, as these constructs are deeply rooted in the psychological theory. Consequently, for better prediction, it becomes imperative to take into account ancillary factors such as the psychological test scores, demographic features, underlying latent primitive emotions, along with the text's undertone and its psychological complexity. This paper proffers team PVG's solution to the WASSA 2021 Shared Task on Predicting Empathy and Emotion in Reaction to News Stories. Leveraging the textual data, demographic features, psychological test score, and the intrinsic interdependencies of primitive emotions and empathy, we propose a multi-input, multi-task framework for the task of empathy score prediction. Here, the empathy score prediction is considered the primary task, while emotion and empathy classification are considered secondary auxiliary tasks. For the distress score prediction task, the system is further boosted by the addition of lexical features. Our submission ranked 1$^{st}$ based on the average correlation (0.545) as well as the distress correlation (0.574), and 2$^{nd}$ for the empathy Pearson correlation (0.517).
\end{abstract}

\section{Introduction}
In recent years, substantial progress has been made in the NLP domain, with sentiment analysis and emotion identification at its core. The advent of attention-based models and complex deep learning architectures has led to substantial headways in sentiment/ emotion classification and their intensity prediction. However, the studies addressing the prediction of affective phenomenons of empathy and distress have been relatively limited. Factors such as lack of large-scale quality labeled datasets, the weak notion of the constructs themselves, and inter-disciplinary dependencies have hindered the progress. The WASSA 2021 Shared Task on Predicting Empathy and Emotion in Reaction to News Stories \citep{tafreshi-etal-2021} provides a quality, gold-standard dataset of the empathic reactions to news stories to predict Batson's empathic concern and personal distress scores.

Empathy, as defined by \citet{davis1983effects}, is considered as \say{reactions of one individual to the observed experiences of another.} It is more succinctly summarized by \citet{levenson1992empathy} in three key components, \say{(a) knowing what another person is feeling (cognitive), (b) feeling what another person is feeling (emotional), and (c) responding compassionately to another person's distress (behavioral).} Distress, on the other hand, as delineated by \citet{dowling2018compassion}, is \say{a strong aversive and self-oriented response to the suffering of others, accompanied by the desire to withdraw from a situation in order to protect oneself from excessive negative feelings.} Empathy and distress are multifaceted interactional processes that are not always self-evident and often depend on the text's undertone. Moreover, along with the textual data, multiple psychological and demographic features also play a vital role in determining these complex emotions. Evidence by \citet{fabi2019empathic} suggests that empathy and distress are not independent of the basic emotions (happiness, sadness, disgust, fear, surprise, and anger) the subject feels during a given scenario. This appositeness of the primitive emotions with empathy and distress can be aptly exploited using a multi-task learning approach. 

Multi-task learning has led to successes in many applications of NLP such as machine translation \citep{mccann2017learned}, speech recognition \citep{arik2017deep}, representation learning \citep{hashimoto-etal-2017-joint}, semantic parsing \citep{peng-etal-2017-deep}, and information retrieval \citep{liu-etal-2015-representation} to name a few. NLP literature \citep{standley2020tasks} suggests that under specific circumstances and with well-crafted tasks, multi-task learning frameworks can often aid models to achieve state-of-the-art performances. \citet{standley2020tasks} further asserts that seemingly related tasks can often have similar underlying dynamics. With the same intuition,  \citet{deep2020related} designed a multi-task learning model for sentiment classification and their corresponding intensity predictions. Building on these findings, we propose a multi-input, multi-task, transformer-based architecture for the prediction of empathy and distress scores.  The multi-input nature of the framework aggregates information from textual, categorical, and numeric data to generate robust representations for the regression task at hand. Exploiting the latent interdependencies between primitive emotions and empathy/ distress, we formulate the multi-task learning problem as a combination of classification and regression. The model simultaneously classifies the text into its correct basic emotion, detects if it exhibits high empathy/ distress, and accordingly predicts its appropriate empathy/ distress intensity score according to Batson's scale \citep{batson1987distress}. This multi-input, multi-task learning paradigm is further bolstered with the addition of NRC Emotion Intensity Lexicons  \citep{mohammad-2018-word}; NRC Valence, Arousal, and Dominance Lexicons \citep{mohammad-2018-obtaining}; and relevant features from Empath \citep{empathcitation}. Moreover, our proposed models have less than 110k trainable parameters and are still able to achieve relatively high Pearson's correlation of 0.517 and 0.574 for empathy and distress, respectively, and 0.545 for average correlation, outperforming other teams.

\section{Related Work}
Over the last few years, earnest endeavours have been made in the NLP community to analyze empathy and distress. Earlier work in empathy mostly addressed the presence or absence of empathy in spoken dialogue \citep{ gibson2015predicting, alam2016can, fung-etal-2016-zara, perez-rosas-etal-2017-understanding, alam2018annotating}. For text-based empathy prediction, \citet{buechel-etal-2018-modeling} laid a firm foundation for predicting Batson's empathic concern and personal distress scores in reaction to news articles. They present the first publicly available gold-standard dataset for text-based empathy and distress prediction. \citet{sharma-etal-2020-computational} contemplated a computational approach for understanding empathy in text-based health support. They developed a multi-task RoBERTa-based bi-encoder paradigm for identifying empathy in conversations and extracting rationales underlying its predictions. \citet{wagner2020s} analysed the linguistic undertones for empathy present in avid fiction readers.

Computational work done for predicting distress is relatively modest. \citet{shapira2020using} analysed textual data to examine associations between linguistic features and clients’ distress during psychotherapy. They combined linguistic features like positive and negative emotion words with psychological measures like Outcome Questionnaire-45 \citep{lambert2004outcome} and Outcome Rating Scale\citep{miller2003outcome}. \citet{zhou-jurgens-2020-condolence} studied the affiliation between distress, condolence, and empathy in online support groups using nested regression models.

\section{Data Description}
The WASSA 2021 Shared Task \citep{tafreshi-etal-2021} provides an extended dataset to the one compiled by \citet{buechel-etal-2018-modeling}. The dataset has a total of 14 features spanning textual, categorical, and numeric data types. Essays represent the subject's empathic reactions to news stories he/she has read. The demographic features of gender, race, education, and the essay's gold-standard emotion label cover the categorical input features. The numeric features include the subject's age and income, followed by personality traits scores (conscientiousness, openness, extraversion, agreeableness, stability) and interpersonal reactivity index (IRI) scores (fantasy, perspective taking, empathetic concern, personal distress). The train-development-test split of the dataset is illustrated in Table~\ref{data-distribution}.

\begin{table}
\centering
\begin{tabular}{ll}
\hline \textbf{Dataset} & \textbf{Datapoints} \\ \hline
Train & 1860 \\
Development & 270 \\
Test & 525 \\
\hline
\end{tabular}
\caption{\label{data-distribution} Data distribution.}
\end{table}

\section{Proposed Methodology}

\begin{figure}[ht]
    \includegraphics[width = \linewidth, height= 8 cm]{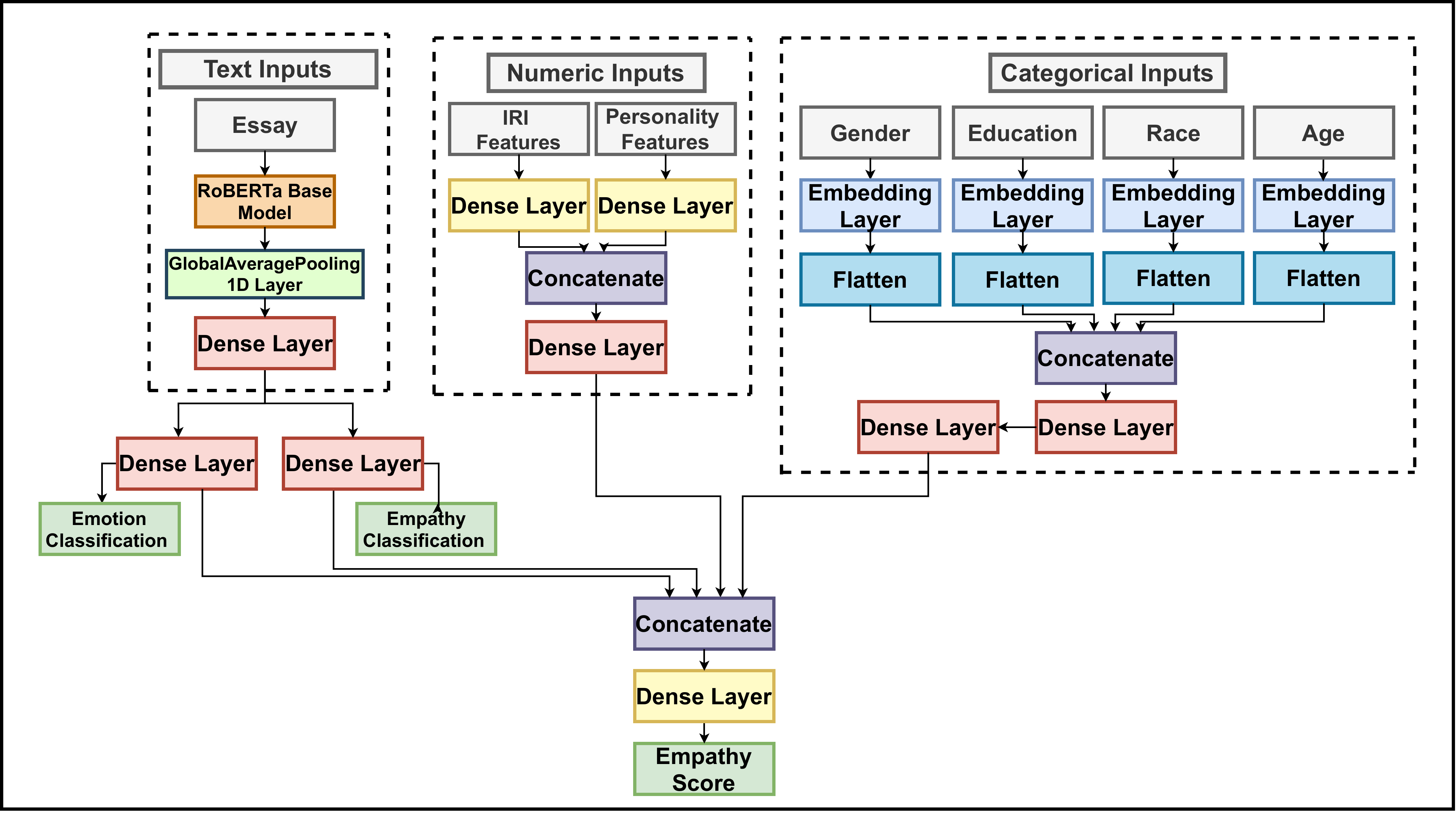}
\caption{\label{model-empathy}System architecture for empathy score prediction. The Dense layers in red have an $\ell^2$ kernel regularization applied to them.}
\end{figure}

\begin{figure}[ht]
    \includegraphics[width = \linewidth, height= 8 cm]{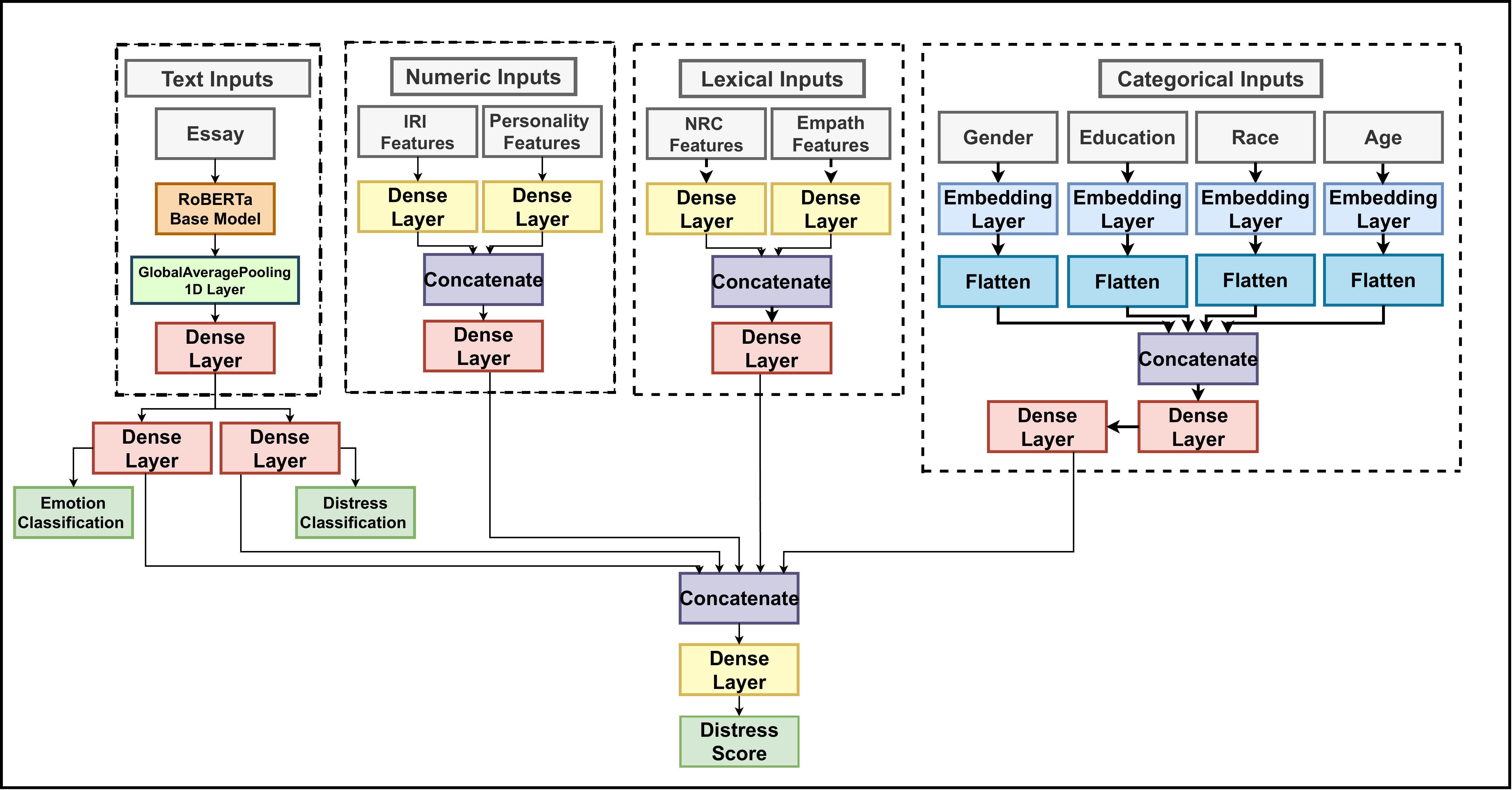}
\caption{\label{model-distress}System architecture for distress score prediction. The Dense layers in red have an $\ell^2$ kernel regularization applied to them.}
\end{figure}

In this section, we posit two multi-task learning frameworks for empathy and distress score prediction. They are elucidated in detail as follows:

\subsection{Empathy Score Prediction}

Figure~\ref{model-empathy} depicts the system architecture for empathy score prediction. We formulate the task of empathy score prediction as a multi-input, multi-task learning problem. The proposed multi-task learning framework aims to leverage the empathy bin \footnote{empathy bin is a feature given in the training dataset, where its value is 1 if empathy score greater than or equal to 4.0, and 0 if empathy score is less than 4.0. Thus, an essay exhibits high empathy if its empathy bin is 1 and exhibits low empathy if its empathy bin is 0. The same analogy is true for distress bin.} and the text's emotion to predict its empathy score. Here, the empathy score prediction is treated as the primary task, whereas emotion and empathy classification are considered secondary auxiliary tasks. The multi-input, multi-task nature of the framework efficiently fuses the diverse set of information (textual, categorical, and numeric) provided in the dataset to generate robust representations for empathy score prediction. 
 
\begin{table*}
\small
\renewcommand{\arraystretch}{1.3}%
\setlength\extrarowheight{2pt} 
\begin{tabularx}{\textwidth}{
  p{\dimexpr.1\linewidth-2\tabcolsep-1.3333\arrayrulewidth}
  p{\dimexpr.45\linewidth-2\tabcolsep-1.3333\arrayrulewidth}
  p{\dimexpr.45\linewidth-2\tabcolsep-1.3333\arrayrulewidth}
  }
\hline
\textbf{} & \textbf{Correlation with empathy score} & \textbf{Correlation with distress score}  \\\hline
\textbf{NRC features} & sadness (0.19), fear (0.12), arousal (0.11),
joy (0.11), valence (0.11), dominance (0.10) & fear (0.22), sadness (0.20), disgust (0.15),
arousal (0.15), anger (0.15), dominance (0.12)\\
\hline
\textbf{Empath features} &  domestic\_work (0.15), death (0.14), home (0.12), sadness (0.12), suffering (0.11), help (0.11), party (0.11), family (0.10), shame (0.10), celebration (0.10), leisure (0.9), negative\_emotion (0.9), air\_travel (0.8), violence (0.8), pain (0.7) &  
suffering (0.16),  death (0.15),  torment (0.13),  hate (0.11),  negative\_emotion (0.10),  sadness (0.10),  aggression (0.09),  fight (0.09),  help (0.09),  pain (0.08),  kill (0.08),  horror (0.08),  violence (0.08),  war (0.08), ugliness (0.7)\\
\hline

\end{tabularx}
\caption{\label{lexical-features} Selected NRC and Empath features and their pearson's correlation with training data's empathy and distress score.}
\end{table*}

For the task of empathy and emotion classification, we make use of the pre-trained RoBERTa base model \citep{liu2019roberta}. The contextualized representations generated by RoBERTa help extract the context-aware information and capture the undertone of the text better than standard deep learning models. For each word in the essay, we extract the default pre-trained embeddings from the last hidden layer of RoBERTa. The 768-dimensional word embeddings are averaged to generate essay-level representations, followed by a hidden layer (128 units) for dimensionality reduction. The model further branches into two parallel fully-connected layers (16 units each), which form the task-specific layers for empathy and emotion classification, respectively. Let $T_1$ and $T_2$ denote the generated task-specific representations for empathy and emotion classification, respectively, of dimension $d_1$. Finally, a classification layer (binary classification for empathy and multi-class classification for emotion) is added to each of the task-specific layers.

To incorporate the demographic information encoded in the categorical variables, we make use of entity embeddings \citep{guo2016entity}.  Formally, entity embeddings are domain-specific multi-dimensional representations of categorical variables, automatically learned by a neural network when trained on a particular task. For each of the demographic features (gender, education, race, and age\footnote{In the dataset age is given as a numeric input. We split it into intervals of below 25, 26-40, 41-60, and 61 and above.}), 3-dimensional entity embeddings are generated. All the resultant embeddings are flattened, concatenated, and passed through two fully-connected layers (32 and 16 units, respectively), generating a layer having relevant information from each categorical input. Let this representation be denoted as $C$ of dimension $d_2$.

The numeric inputs of the personality and the Interpersonal Reactivity Index (IRI) scores are incorporated in the model by passing them individually through a single hidden layer (8 units). The results are concatenated and further passed on to a fully-connected layer (32 units) to generate their combined representations. Let this representation be denoted as $N$ of dimension $d_3$.

The task-specific layers for empathy and emotion classification and the representations generated by the final hidden layer of the entity embeddings and the numeric psychological score inputs are concatenated as given in equation~\ref{concat-empathy} to generate the final representation $F_1$ for empathy score prediction.
\begin{equation}
\label{concat-empathy}
F_1 = [T_1; T_2; C; N]  \in \mathbb{R}^{d_1+d_2+d_3}
\end{equation}

It is further passed to another hidden layer (16 units). Thus, this layer contains the compressed information from different knowledge views of the input data. A final regression layer is added for the empathy score prediction task. The multi-input, multi-task model is trained end-to-end with the objective loss calculated as the sum of the loss for each of the three tasks.

\subsection{Distress Score Prediction}

Figure~\ref{model-distress} depicts the system architecture for distress score prediction. The distress score prediction model is the same as that of the empathy score prediction model, but with the addition of hand-crafted lexical features. We use 6 NRC Emotion Intensity Lexicons  \citep{mohammad-2018-word}; 2 NRC Valence, Arousal, and Dominance Lexicons \citep{mohammad-2018-obtaining}; and 15 features from Empath \citep{empathcitation} as stated in Table ~\ref{lexical-features}. These features are chosen as they exhibit a high Pearson correlation with the training data. For each essay, the respective NRC and Empath score is calculated as the sum of each word's score in the essay. The NRC lexicons and Empath features are passed to a single hidden layer (8 and 16 units for NRC and Empath, respectively.) independently before concatenation. The resultant representation is further passed through another hidden layer (48 units). Let this representation be denoted by $L$ of dimension $d_4$. It is concatenated with the final layer representations from other inputs to generate the final representation $F_2$ for distress score prediction, as given by equation~\ref{concat-distress}.
\begin{equation}
\label{concat-distress}
F_2 = [T_1; T_2; C; N; L]  \in \mathbb{R}^{d_1+d_2+d_3+d_4}
\end{equation}

\section{Experimental Setup}

\subsection{Data Preparation}
We applied standard text cleaning steps for each essay in the dataset, such as removing the punctuations, special characters, digits, single characters, multiple spaces, and accented words. The essays are further normalized by removing wordplay, replacing acronyms with full forms, and expanding contractions \footnote{\url{https://pypi.org/project/pycontractions/}}. 

Each essay is tokenized and padded to a maximum length of 200 tokens. Longer essays are truncated. Each Empath feature is converted into its percentage value. For distress score prediction, the numeric features, NRC lexicon scores, and Empath features are standardized by removing the mean and scaling to unit variance before being passed to the model.

\begin{table}
\centering
\begin{adjustbox}{width=\columnwidth}
\begin{tabular}{l l l l l }
    \hline
        \multirow{3}{*}{\textbf{Model Name}}
       
        & \multicolumn{2}{c}{\textbf{Empathy}}
        & \multicolumn{2}{c}{\textbf{Distress}} \\
     
  &   \textbf{Train}&\textbf{Dev}&\textbf{Train}&\textbf{Dev}   \\
    \hline Plain RoBERTa model  &0.56&0.47&0.56&0.41  \\
Plain multi-input model   & 0.64 & 0.50 & 0.64 & 0.49   \\
\textbf{Multi-input, multi-task} \\  \textbf{model} & \textbf{0.74}  & \textbf{0.57}  &  0.66 & 0.53    \\
\textbf{Multi-input, multi-task} \\ \textbf{model+ NRC + Empath} & 0.64  & 0.54  & \textbf{0.66} & \textbf{0.56}  \\
\hline
\end{tabular}
\end{adjustbox}
\caption{\label{results} Performance comparison of various models as per Pearson's correlation ($p$ $<$ 0.05).}
\end{table}

\subsection{Parameter Setting and Training Environment}
Given the small amount of data, the weights of the RoBERTa layers were freezed and not updated during the training. The multi-task model is trained end-to-end using the Adam optimizer \citep{kingma2014adam} with a learning rate of \num{1e-3} and a batch size of 32. We used Hyperbolic Tangent (tanh) activation for all the hidden layers as it performed better than ReLU activation and its other variants. The model is trained for 200 epochs with early stopping applied if the validation loss does not improve after 20 epochs. Furthermore, the learning rate is reduced by a factor of 0.2 if validation loss does not decline after ten successive epochs. The model with the best validation loss is selected. $\ell^2$ kernel regularizer of $\num{5e-4}$ applied to certain hidden layers as shown in Figures~\ref{model-empathy} and~\ref{model-distress}. A dropout of probability 0.2 is applied after the average pooling of contextual RoBERTa embeddings. Our code is available at our GitHub repository.\footnote{\url{https://github.com/mr-atharva-kulkarni/WASSA-2021-Shared-Task}}

\section{Results and Discussion}
The performance comparison of the various models on the validation data is reported in Table~\ref{results}. As illustrated in the Table~\ref{results}, we compare the performances of a RoBERTa based regression model, a multi-input model (text + categorical + numeric); a multi-input (text + categorical + numeric), multi-task model;  and a multi-input (text + categorical + numeric), multi-task model with added lexical features (NRC + Empath). The systems submitted by our team are highlighted in Table~\ref{results}. It is evident from Table~\ref{results} that the addition of categorical and numeric input features leads to an appreciable improvement in the model’s performance. This further attests that the demographic features and the psychological scores contribute valuable information for predicting scores of complex emotions like empathy and distress. The performance is further improved by adopting a multi-task approach, thus, reinstating our belief in the interdependencies between the primitive emotions and the complex emotions of empathy and distress. The addition of lexical features, however, leads to an improvement for distress prediction but a decrease in performance for empathy prediction. This may be explained by the theories in the affective neuroscience literature, wherein empathy is considered a neocortex emotion, describing it as an emotional, cognitive, and behavioral process. Thus, one’s empathic quotient is reflected in how one expresses one’s feelings and the undertone of it, rather than the mere words one uses. Moreover, as stated by \citet{sedoc-etal-2020-learning} in their work, there exist no clear set of lexicons that can accurately distinguish empathy from self-focused distress. Another reason for the decrease in the performance for empathy score prediction may be attributed to the fact that the Empath features for empathy, as reported in Table~\ref{lexical-features} lack some interpretability from a human perspective. Empath features such as \textit{domestic work, party, celebration, leisure,} and \textit{air travel} are not innately empathic categories and perhaps show high correlation due to corpus-based topical bias. The Empath features for distress, on the other hand, seem quite relevant and thus, might explain the increase in performance for distress prediction. We encourage further research in this direction.

\section{Conclusion and Future Work}
In this work, we propose a multi-input, multi-task, transformer-based architecture to predict Batson's empathic concern and personal distress scores. Leveraging the dependency between the basic emotions and empathy/ distress, as well as incorporating textual, categorical, and numeric data, our proposed model generates robust representations for the regression task at hand. The addition of certain lexical features further improves the model's performance for distress score prediction. Our submission ranked 1$^{st}$ based on average correlation (0.545) as well as distress correlation (0.574), and 2$^{nd}$ for empathy Pearson correlation (0.517) on the test data. As for the future work of this research, a weighted loss scheme could be employed to enhance the results. From a psychological and linguistic standpoint, features such as part-of-speech tags, syntax parse tree, and the text's subjectivity and polarity scores could also be exploited.

\bibliography{eacl2021}
\bibliographystyle{acl_natbib}

\end{document}